\def\year{2021}\relax
\documentclass[letterpaper]{article} 
\usepackage{aaai21}  
\usepackage{times}  
\usepackage{helvet} 
\usepackage{courier}  
\usepackage[hyphens]{url}  
\usepackage{graphicx} 
\usepackage{natbib}  
\usepackage{caption} 
\usepackage{multirow}
\usepackage{amsthm,amsmath,amssymb}
\usepackage{mathrsfs}
\usepackage{color}
\usepackage{pifont}
\usepackage{amssymb}
\usepackage[switch]{lineno}
\title{MANGO: A Mask Attention Guided One-Stage Scene Text Spotter}
\author{Liang Qiao\textsuperscript{\rm 1}\footnotemark[1],
        Ying Chen\textsuperscript{\rm 2}\thanks{Authors contribute equally. Chen did this work during an internship in Hikvision Research Institute.},
        Zhanzhan Cheng\textsuperscript{\rm 31}\thanks{This work is completed under the supervision of Zhanzhan Cheng (contact email:
chengzhanzhan@hikvision.com).},
        Xunlu Xu\textsuperscript{\rm 1},
        Yi Niu\textsuperscript{\rm 1},
        Shiliang Pu\textsuperscript{\rm 1}\thanks{Corresponding author},
        Fei Wu\textsuperscript{\rm 3}\\
}
\affiliations{
\textsuperscript{\rm 1}Hikvision Research Institute, China;
~~~\textsuperscript{\rm 2}Tongji University, China;
~~~\textsuperscript{\rm 3}Zhejiang University, China\\
\{qiaoliang6,~chengzhanzhan,~xuyunlu,~niuyi,~pushiliang\}@hikvision.com~~1833019@tongji.edu.cn~~wufei@cs.zju.edu.cn
}
 
\begin{document}

\maketitle
\begin{abstract}
Recently end-to-end scene text spotting has become a popular research topic due to its advantages of global optimization and high maintainability in real applications.
Most methods attempt to develop various region of interest (RoI) operations to concatenate the detection part and the sequence recognition part into a two-stage text spotting framework.
However, in such framework, the recognition part is highly sensitive to the detected results (\emph{e.g.}, the compactness of text contours).
To address this problem, in this paper, we propose a novel Mask AttentioN Guided One-stage text spotting framework named MANGO, in which character sequences can be directly recognized without RoI operation.
Concretely, a position-aware mask attention module is developed to generate attention weights on each text instance and its characters.
It allows different text instances in an image to be allocated on different feature map channels which are further grouped as a batch of instance features.
Finally, a lightweight sequence decoder is applied to generate the character sequences.
It is worth noting that MANGO inherently adapts to arbitrary-shaped text spotting and can be trained end-to-end with only coarse position information (\emph{e.g.}, rectangular bounding box) and text annotations.
Experimental results show that the proposed method achieves competitive and even new state-of-the-art performance on both regular and irregular text spotting benchmarks, i.e., ICDAR 2013, ICDAR 2015, Total-Text, and SCUT-CTW1500.
\end{abstract}

\section{Introduction}
Scene text spotting has attracted much attention due to its various practical applications such as key entities recognition in invoice/receipt understanding, product name identification in the e-commerce system, and license plate recognition in the intelligent transportation system.
Traditional scene text spotting systems are usually in three steps: localizing text regions, cropping text regions from the original image, and recognizing them as character sequences \cite{wang2012end,jaderberg2014deep,neumann2015real,gomez2017textproposals}.
While such text spotting model brings many considerable problems, such as (1) errors will be accumulated among the multiple individual tasks, (2) it is costly to maintain multiple separate models, and (3) the model is hard to adapt to various applications.
\begin{figure}[!t]
\begin{center}
\includegraphics[width=1.0\linewidth]{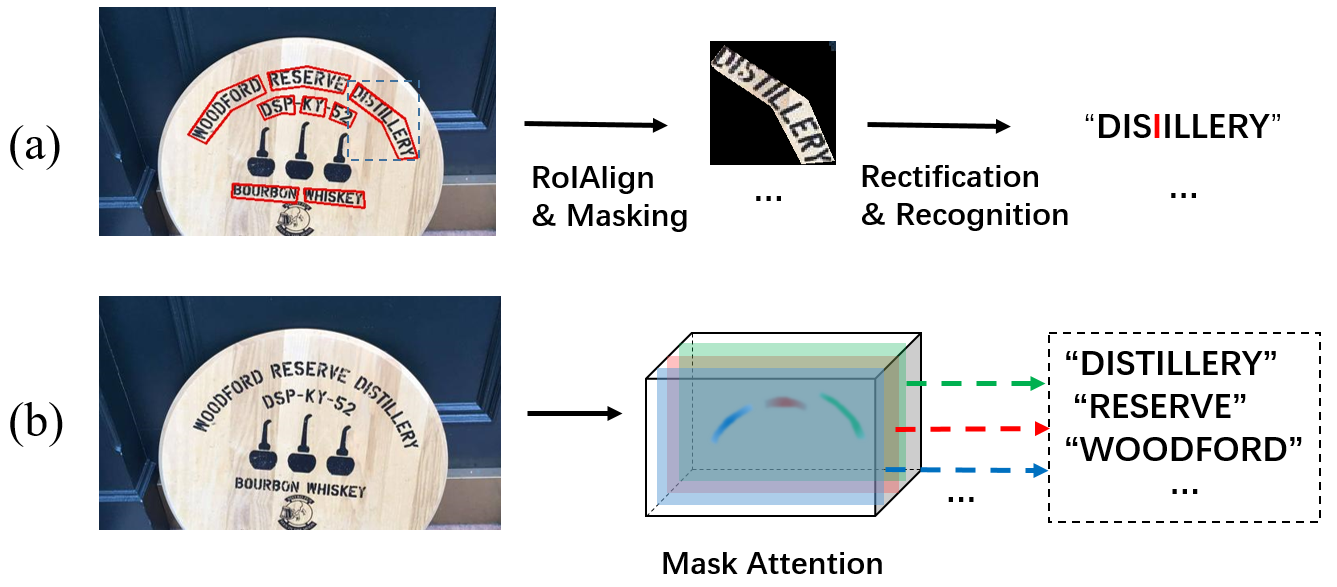}\\
\end{center}
\caption{
Illustration of the traditional two-stage text spotting process and the proposed MANGO.
Sub-figure (a) shows the two-stage text spotting strategy by using RoI operations to connect the detection and recognition parts.
Sub-figure (b) is the proposed one-stage text spotting approach, which can directly output the final character sequences.
}
\label{fig:0}
\end{figure}

Therefore, many works \cite{li2017towards,bartz2017see,he2018end,sun2018textnet,liao2019mask} are proposed to optimize the text spotting process in an end-to-end manner.
These methods usually use various Region of Interest (RoI) operations to bridge the text detection and recognition parts in a differentiable way, which form the \emph{two-stage} framework.
Roughly speaking,
the early end-to-end methods \cite{li2017towards,busta2017deep,liu2018fots,he2018end} used the axis-aligned  rectangular RoIs as the connecting modules. These methods are limited to cope with irregular (\emph{e.g.}, perspective, or curved) text instances since such kind of RoIs might bring interferences from background or other texts.
To solve this problem, the later methods \cite{feng2019textdragon,qiao2020text,wang2020all,qin2019towards,liu2020abcnet} designed some shape-adaptive RoI mechanisms to extract the irregular text instances and rectify them into regular shapes.

In \emph{two-stage} methods, the recognition part highly depends on the localization results, which needs the detection part must be capable of capturing accurate text boundaries to eliminate the background interference.
Thus, training a robust text detection model relies on accurate detection annotations, such as polygonal or mask annotations used in irregular text spotting.
Naturally, labeling such kind of annotations is laborious and costly.
On the other hand, it is not easy to ensure that the tightly enclosed text regions (supervised by detection annotations) are the best form for the following recognition task. For example, in Figure \ref{fig:0}(a), tight text boundaries may erase the edge texture of characters and lead to erroneous results.
These tight detection results often need to be expanded manually for adapting to recognition well in real applications.
Besides, performing complex RoI operations with non-maximum suppression (NMS) after proposals is also time-consuming, especially for arbitrary-shaped regions.
Though \cite{xing2019convolutional} proposed a \emph{one-stage} character-level spotting framework with its character segmentation strategy,
it is difficult to extend to the situations with more character classes (\emph{e.g.}, Chinese characters). It also loses crucial context information among  characters.

In fact, when people read, they do not need to depict the accurate contours of text instances. It is enough to identify text instance via rough text position attended by visual attention.
Here, we rethink the scene text spotting as an attending and reading problem, i.e., directly reading out the text contents of the coarsely attended text regions all at once.

In this paper, we propose a Mask Attention Guided One-stage text spotter called MANGO, a compact and powerful \emph{one-stage} framework that directly predicts all texts simultaneously from an image without any RoI operation.
Specifically, we introduce a position-aware mask attention (PMA) module to generate spatial attention over text regions, which contains both the instance-level mask attention (IMA) part and the character-level mask attention (CMA) part.
IMA and CMA are responsible for perceiving the positions of text and characters in an image, respectively.
Text instances' features can be directly extracted by the position-aware attention maps rather than explicit cropping operation, which reserves the global spatial information as much as possible.
Here, different text instances' features will be mapped into different feature map channels using dynamic convolutions \cite{wang2020solov2}, as shown in Figure \ref{fig:0}(b).
After that, a lightweight sequence decoder is applied to generate character sequences in a batch all at once.

Note that MANGO can be end-to-end optimized with only rough position information (\emph{e.g.}, a rectangular bounding box, or even the center point of the text instance) as well as sequence annotations.
Benefiting from PMA, this framework can adaptively spot various irregular text without any rectification mechanism, and is also capable of learning the reading order for arbitrary-shaped text.

The major contributions of this paper are as follows:
(1) We propose a compact and robust \emph{one-stage} text spotting framework named MANGO that can be trained in an end-to-end manner.
(2) We develop the position-aware mask attention module to generate the text instance features into a batch, and build the one-to-one mapping with final character sequences.
The module can be trained with only rough text position information and text annotations.
(3) Extensive experiments show that our method achieves competitive and even state-of-the-art results on both regular and irregular text benchmarks.

\section{Related Works}
We divide existing scene text spotting methods into the following two categories.
\begin{figure*}[th!]
\begin{center}
\includegraphics[width=0.83\textwidth, height=6cm]{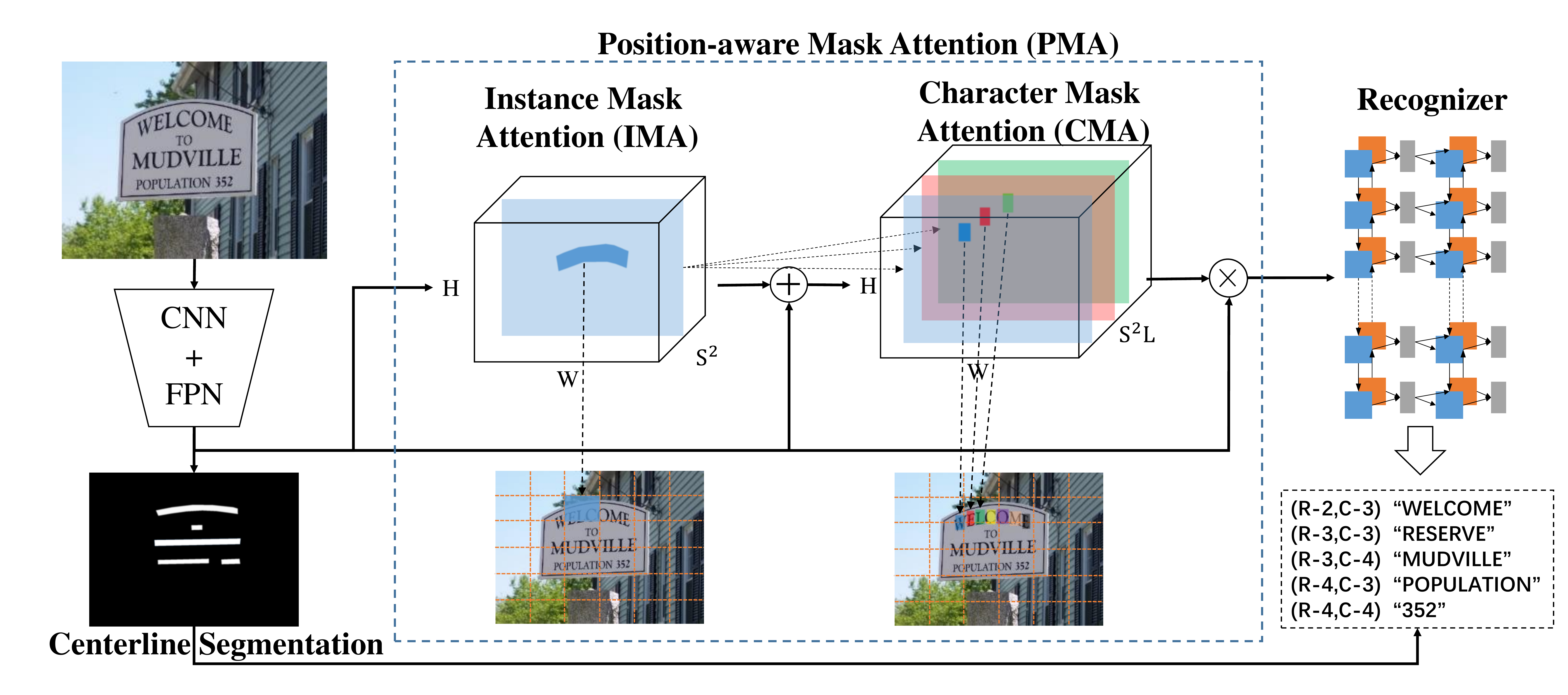}\\
\end{center}
\caption{The workflow of MANGO. We take $S$$=$$6$ as an example. The input features are fed into a Position-aware Mask Attention module to map different features of instances/characters into different channels. The recognizor finally outputs character sequences in a batch all at once. A Centerline Segmentation branch is used to generate the rough positions of all text instances. Prefix `R-' and `C-' separately denote the grid row and column.}
\label{framework}
\end{figure*}
\subsection{Two-stage End-to-end Scene Text Spotting}
Early scene text spotting methods \cite{liao2018textboxess,liao2017textboxes,wang2012end} usually first localize each text with a trained detector such as \cite{liao2017textboxes,zhou2017east,he2017single,ma2018arbitrary,xu2019textfield,baek2019character} and then recognize the cropped text region with a sequence decoder \cite{shi2016robust,shi2017end,cheng2017focus,zhan2019esir,luo2019moran}.
To sufficiently exploit the complementarity between text detection and text recognition, some works have been proposed to optimize the scene text spotting framework in an end-to-end manner, in which module connectors (\emph{e.g.}, RoI Pooling \cite{ren2015faster} used in \cite{li2017towards,wang2019towards}, RoI-Align used in \cite{he2018end} and RoI-Rotate used in \cite{liu2018fots}) are developed to bridge the text detection and text recognition parts.
Notice that these methods are incapable of spotting arbitrarily shaped text.

To address the irregular problems, many recent works have been proposed to design various adaptive RoI operations to spot arbitrary-shape text.
\cite{sun2018textnet} adopted a perspective RoI transforming module to rectify perspective text, but this strategy still has difficulty in handling heavily curved text.
\cite{liao2019mask} proposed the mask textspotter inspired by the two-stage Mask-RCNN for detecting arbitrarily shaped text character-by-character, but this method loses the context information of characters and requires character-level location annotations.
\cite{qin2019towards} directly adopted Mask-RCNN and an attention-based text recognizer using an RoI-Masking module to remove the background interferences before recognition.
\cite{feng2019textdragon} treated a text instance as a group of feature pieces and adopted the RoI-Slide operation to reconstruct a straight feature map.
Both \cite{qiao2020text} and \cite{wang2020all} detected the key points around text and applied the thin-plate-spline transformation \cite{bookstein1989principal} to rectify irregular instances.
To obtain the smooth feature of the curved text, \cite{liu2020abcnet} used a Bezier curve to represent the top and bottom boundaries of text instances, and proposed a Bezier-Align operation to obtain the rectified feature maps.

The above methods achieve the end-to-end scene text spotting in a two-stage framework, in which the RoI-based connectors (\emph{e.g.}, RoI-Align, RoI-Slide and Bezier-Align, etc.) need to be designed to explicitly crop the feature map.
In two-stage frameworks, the performance highly depends on the text boundary accuracy acquired by the RoI operations. However, these complicated polygonal annotations are usually expensive and not always suited to the recognition part, as mentioned previously.

\subsection{One-stage End-to-end Scene Text Spotting}
In general object localization area, many recent advances have demonstrated the efficiency and effectiveness of one-stage frameworks studied in Object Detection \cite{Redmon2016You, liu2016ssd,lin2017focal,tian2019fcos, duan2019centernet} or Instance Segmentation \cite{wang2019solo,tian2020conditional,wang2020solov2,xie2020polarmask,chen2020blendmask}.
However, scene text spotting is a much more challenging task, since it involves the sequence recognition problem instead of single object classification.
This is because scene text has many particular traits: arbitrary shaped (e.g., curve, slant or perspective, etc.), millions of character combinations, and even unconstrained reading orders (e.g., from right to left).
Recently, \cite{xing2019convolutional} proposed a one-stage scene text spotting approach by directly segmenting single characters. However, it loses the sequence context information among individual characters and is hard to be transferred to more character classes.
To the best of our knowledge, there is no previous work to cope with the sequence-level scene text spotting task in a one-stage framework.

\section{Methodology}
\subsection{Overview}
We propose a one-stage scene text spotter named MANGO, as shown in Figure \ref{framework}.
The deep features are extracted through the backbone of ResNet-50 \cite{he2016deep} and a feature pyramid network (FPN) \cite{lin2017feature}.
The generated feature maps are then fed into three learnable modules:
(1) The position-aware mask attention (PMA) module for learning spatial attention of individual text instances, which consists of the instance-level mask attention (IMA) sub-module and the character-level mask attention (CMA) sub-module.
(2) The sequence decoding task for decoding the attending instance features as character sequences.
(3) The global text centerline segmentation task for providing the rough text position information in the inference stage.
\subsection{Position-aware Mask Attention Module}

A one-stage text spotting problem can be treated as a pure text recognition task in the original image. The critical step is to build the direct one-to-one mapping between the text instances to the final character sequences in a fixed order.
Here, we develop the position-aware attention (PMA) module to capture all represented text features once for the following sequence decoding module.
Inspired by the grid mapping strategy used in \cite{wang2019solo}, we find that different instances can be mapped into different specific channels and achieve the instance-to-feature mapping.
That is, we first divide the input image into $S$$\times$$S$ grids. Then the information around a grid will be mapped into the specific channel of feature maps by the proposed PMA module.

Specifically,
we denote the obtained feature map after feature extraction as $x$$\in$$\mathbb{R}^{C\times H\times W}$, where $C$, $H$ and $W$ are channel size, width and height of the feature map, respectively.
We then feed $x$ into PMA (including IMA and CMA modules) to generate the feature representations of text instances (described bellow).

\subsubsection{Instance-level Mask Attention}
IMA is responsible for generating the instance-level attention mask and assigning different instances' features into different feature map channels. It is achieved by operating a group of dynamic convolutional kernels \cite{wang2020solov2} on the sliced grids, denoted as $G^{S\times S\times C}$. The kernel size is set as $1$$\times$$1$.

Therefore, the instance-level attention mask can be generated by applying these kernels to the original feature map:
\begin{equation}
x_{ins} = G(x),
\label{eq-1}
\end{equation}
where $x_{ins} \in \mathbb{R}^{S^2 \times H \times W}$.
Note that the generated feature channels are corresponding to the grid numbers.

To learn the dynamic convolutional kernels $G$, we need to make grid matching between the text instances and grids.
Unlike general object detection or instance segmentation task, text instances usually appear in a large aspect ratio or even seriously curved.
It is not reasonable to directly use the center of the text bounding box to perform grid matching.
Hence, we define the term \emph{occupation ratio} $o_{i,j}$ to represent how closely a text instance $t_i$ matches a grid $g_j$:
\begin{equation}
o_{i,j}=\max\left(\frac{Inter(A(g_j),A(t_i))}{A(g_j)}, \frac{Inter(A(g_j),A(t_i))}{A(t_i)} \right),
\end{equation}
where $A(.)$ is the region area and $Inter(.,.)$ is the intersection area of two regions.
We say that text instance $t_i$ \emph{occupies} a grid $g_j$ if $o_{i,j}$ is larger than a preset threshold $\mu$.
Then the feature channel $j$ of $x_{ins}$ is in charge of learning the attention mask of text $t_{i}$.
In our experiments, $\mu$ is set to $0.3$.
Note that, in the training stage, \emph{occupation ratio} is calculated based on the shrunk detected ground truth, such as text centerline regions.

For example in Figure \ref{framework}, we set S=6. The word `WELCOME' occupies the (row-2, col-3) and (row-2,col-4) grids. Thus, the 9-th ($(2-1)\times 6+3$) and the 10-th ($(2-1)\times 6 + 4$) grids will predict the same attention mask.
If there are two instances \emph{occupying} the same grid, we simply choose the one with a larger \emph{occupation ratio}.

\subsubsection{Character-level Mask Attention}
As many works \cite{cheng2017focus,xing2019convolutional} demonstrated, the character-level position information can help to improve the recognition performance.
This inspires us to design the global character-level attention submodule to provide the fine-grained feature for the subsequent recognition task.

As shown in Figure \ref{framework}, CMA first concatenates the original feature map $x$ and the instance-level attention mask $x_{ins}$, and then two convolutional layers (kernel size=$3$$\times$$3$) are followed to predict the character-level attention mask:
\begin{equation}
x_{char} = f(x_{ins}\oplus x),
\end{equation}
where $x_{char} \in \mathbb{R}^{(S^2\times L) \times H \times W}$ and $\oplus$ means the channel-wise concatenation.
Here, $L$ is the predefined maximum length of character strings.

With the same grid matching strategy to IMA, if a text instance $t_i$ occupies grid $g_j$ at (row-$h$,col-$w$), the $((h$$-$$1)$$\times$$S$$\times$$L$$+$$(w$$-$$1)$$\times$$L$$+$$k)$ channel of $x_{char}$ is in charge of predicting the text's $k$-th character mask.
We again take the word `WELCOME' as an example (See Figure  \ref{framework}).
If $L=25$, then the 151-st $((2-1)\times 6\times 25 + (3-1)\times 25 + 1)$ channel predicts the attention mask of the character `W', and the 152-nd channel predicts `E' and so on.

\subsection{Sequence Decoding Module}
Since attention masks of different text instances are allocated to different feature channels, we can packet the text instance features into a batch.
A simple idea is to conduct the attention fusion operation as used in \cite{wang2020decoupled} to generate the batched sequential features $x_{seq}$, i.e.,
\begin{equation}
x_{seq} = x'_{char} \otimes x'^\top,
\label{eq1}
\end{equation}
where $x_{seq} \in \mathbb{R}^{S^2\times L \times C}$,  $\otimes$ is the matrix multiplication operation. $x'_{char}\in \mathbb{R}^{(S^2\times L) \times (H \times W)}$ and $x'\in \mathbb{R}^{C \times (H \times W)}$ are reshaped matrices of $x_{char}$ and $x$, respectively.

Then we can transfer the text spotting problem as a pure sequence classification problem.
The following sequence decoding network is responsible for generating a batch ($S^2$) of character sequences.
Concretely, we add two layers of Bidirectional long short-term memory (BiLSTM) \cite{hochreiter1997long} on $x_{seq}$ to capture the sequential relations, and finally output the character sequences by a fully connected (FC) layer.
\begin{equation}
x_{recog} = FC(BiLSTM(x_{seq}))
\label{e-recog}
\end{equation}
where $x_{recog}\in \mathbb{R}^{S^2\times L \times M}$ and $M$ is the size of character dictionary (including 26 letters, 10 digits, 32 ASCII punctuation marks and 1 EOS symbol).
In specific, if the length of the predicted character string is less than $L$, the rest of the predictions are supplemented with the EOS symbols.

Since $x_{ins}$ are sparse at most time, we only focus on the positive ($o_{i,j}$$>$$\mu$) samples in $x_{ins}$ for reducing computational cost. In both training and inference stages, after the computation of Equation (\ref{eq-1}), we dynamically choose positive channels of the feature map as follows:
\begin{equation}
x'_{ins} = \oplus_{\substack{j \in S^2, \\ t_i \in \mathcal{T}, \\ o_{i,j}>\mu}} x_{ins}[j],
\label{eq0}
\end{equation}
where $x'_{ins} \in \mathbb{R}^{N\times H \times W}$, $x_{ins}[j]$ denotes the $j$-th channel of $x_{ins}$ and $\mathcal{T}$ is the set of text instances. $N$ is the dynamic selected number, which equals to the number of grids that are \emph{occupied} by texts.
Then, $x_{mul}$ in Equation (\ref{eq1}) and $x_{recog}$ in Equation (\ref{e-recog}) can be separately rewritten as $x_{seq} \in \mathbb{R}^{N\times L \times C}$ and $x_{recog} \in \mathbb{R}^{N\times L \times M}$.

\subsection{Text Centerline Segmentation}
The model is now able to output all predicted sequences for $S^2$ grids separately. However, if there are more than two text instances in an image, we still need to point out which grids correspond to those recognition results. Therefore, a text detection branch is required.

Since our method does not rely on the accurate boundary information, we can apply any text detection strategy (\emph{e.g.}, RPN \cite{2015Faster} and YOLO \cite{Redmon2016You}) to obtain the rough geometry information of text instances.
Considering that scene texts might be arbitrary-shaped, we follow most segmentation-based text detection methods \cite{long2018textsnake, Wang2019Shape} to learn the global text centerline region segmentation (or shrunk ground truth) for individual text instances.

\subsection{Optimization}
Both IMA and CMA modules serve to make the network focus on the specific instance and character positions, which can be learned theoretically by only the final recognition part.
However, in complicated scene text scenarios,  it might be difficult for the network to converge without the assistance of position information.
Nevertheless, we find that the model can be easily transferred if it has been pre-trained on the synthetic datasets with character-level supervision in advance. Therefore, the model can be optimized in two steps.

First, we can treat the learning of IMA and CMA as pure segmentation tasks. Together with centerline region segmentation, all segmentation tasks are trained using binary Dice coefficient loss \cite{milletari2016v}, and the recognition tasks simply use cross-entropy loss.
The global optimization can be written as
\begin{equation}
\mathcal{L} = \lambda_{1}\mathcal{L}_{cls} + \lambda_{2}\mathcal{L}_{I}+\lambda_{3}\mathcal{L}_{C} + \mathcal{L}_{recog},
\end{equation}
where $\mathcal{L}_{cls}$, $\mathcal{L}_{I}$, $\mathcal{L}_{C}$ and $\mathcal{L}_{recog}$ denote the losses generated by the centerline segmentation, IMA, CMA and recognition, respectively. $\lambda_{1},\lambda_{2}$ and $\lambda_{3}$ are weighted parameters.

Given the pre-trained weights on synthetic datasets, the model can be simply optimized at any scene by
\begin{equation}
\mathcal{L} = \lambda\mathcal{L}_{cls} + \mathcal{L}_{recog}.
\end{equation}

Note that the pre-training step is actually a one-off task, and the CMA and IMA will then be mainly learned to fit the recognition task. In contrast to previous methods that need to balance the weights of detection and recognition, the end-to-end results of MANGO are mostly supervised by the final recognition tasks.

\subsection{Inference}
In the inference stage, the network outputs a batch ($S\times S$) of probability matrices ($L\times M$). According to the predictions of the centerline segmentation task, we can determine which grids should be treated as valid.
\begin{figure}
\begin{center}
\includegraphics[width=0.8\linewidth]{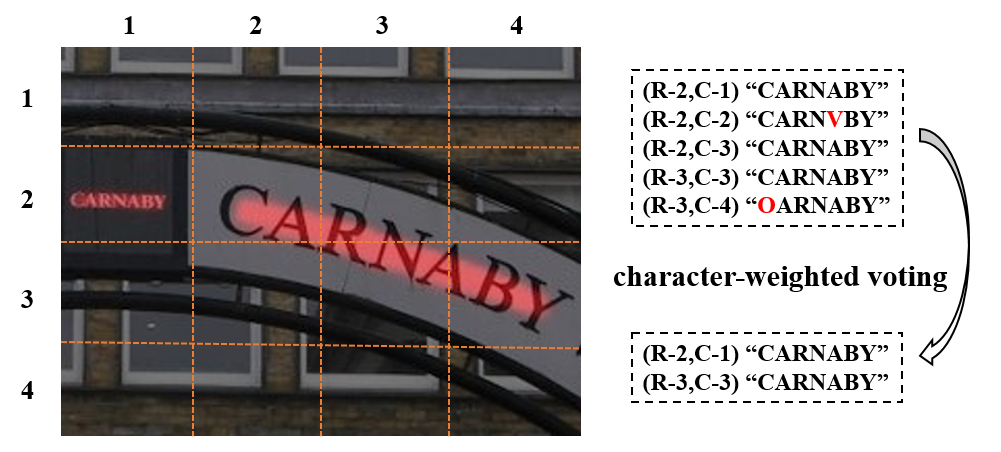}\\
\end{center}
\caption{
Illustration of the inference process. The final predictions are generated by merging all \emph{occupied} grids' results through the character-weighted voting strategy.
}
\label{fig:inference}
\end{figure}
We first conduct a Breadth-First-Search (BFS) to find the individual connected regions.
Many text-like textures can be filtered during this process.
Since each connected region may intersect with several grids, we adopt a character-weighted voting strategy to generate the final character strings, as shown in Figure \ref{fig:inference}.

Specifically, we calculate the \emph{occupation ratio} $o_{i,j}$ of the connected region $i$ with the grid $j$ as the weight of each character.
For the $k$-th character of the instance $i$, its character-weighted voting result is achieved by
\begin{equation}
instance_{i}^{(k)} = \arg\max\left(\sum_{j \in (S\times S)} (o_{i,j}\cdot x_{recog}[j][k])\right),
\end{equation}
where $x_{recog}(j,k)\in \mathbb{R}^{M}$ is the predicted probability vector of the $k$-th character of the $j$-th grid.
Here, the \emph{occupation ratio} provides the confidence of each grid, and multiple outputs fusion could generate more reliable results. The grid with the maximum occupation ratio will be treated as the rough output position, which can be replaced by any form according to the specific task.

\section{Experiments}
\subsection{Datasets}
We list the datasets used in this paper as follows.

\textbf{Training Data.}
We use \emph{SynthText 800k} \cite{gupta2016synthetic} as the pretraining dataset. Both instance-level and character-level annotations are exploited to pre-train the PMA module.

In the finetuning stage, we aim to obtain a general text spotter supporting both regular and irregular scene text reading. Here, we construct a general dataset used for finetuning, which includes 150k images from the Curved SynthText \cite{liu2020abcnet}, 13k images filtered from COCO-Text \cite{2016COCO}, 7k images filtered from ICDAR-MLT \cite{2019ICDAR2019} as well as all training images in ICDAR2013 \cite{karatzas2013icdar}, ICDAR2015 \cite{karatzas2015icdar} and Total-Text \cite{ch2017total}.
Note that, here we use only the instance-level annotations to train the network.

\textbf{Testing Dataset.}
We evaluate our method on two standard text spotting benchmarks \emph{ICDAR2013} \cite{karatzas2013icdar} (IC13)  and \emph{ICDAR2015} \cite{karatzas2015icdar} (IC15), which mainly contain horizontal and perspective text, and two irregular benchmarks \emph{Total-Text} \cite{ch2017total} and \emph{SCUT-CTW1500} \cite{liu2019curved} (CTW1500), which contains many curved text.

We also demonstrate the generalization ability of our method in a license plate recognition dataset, CCPD \cite{xu2018towards}.

\begin{table*}[thb!]
\begin{center}
\scalebox{0.8}{
\begin{tabular}{|c|l|c|c|c|c|c|c|c|c|c|c|c|c|}
\hline
\multirow{2}{*}{Dataset} & \multirow{2}{*}{Method}  &\multirow{2}{*}{Input Size} & \multicolumn{3}{c|}{End-to-End} & \multicolumn{3}{c|}{Word Spotting} & \multirow{2}{*}{FPS}\\ \cline{4-9}
 & && S & W & G & S & W & G &  \\
\hline
\multirow{9}{*}{IC13}
&He et al. $\dagger$ \cite{he2018end}  & -  & 91.0 & 89.0 & 86.0 & 93.0 & 92.0 & 87.0 & -\\
&FOTS $\dagger$ \cite{liu2018fots} & L-920 & 88.8 & 87.1 & 80.8 & 92.7 & 90.7 & 83.5 &  \textbf{22.0}\\
&TextNet \cite{sun2018textnet} & L-920  & 89.8 & 88.9 & 83.0 & 94.6 & \textbf{94.5} & 87.0 & 2.7\\
&Mask TextSpotter$^*$ \cite{liao2019mask}  & S-1000 &  93.3 &91.3 & 88.2 & 92.7 & 91.7 & {87.7} &  3.1 \\
&Boundary \cite{wang2020all}  & L-1280  &  {88.2} & {87.7} & {84.1} & - & - & - &  - \\
&Text Perceptron \cite{qiao2020text} & L-1440  &  {91.4} & {90.7} & {85.8} & \textbf{94.9}& 94.0 & 88.5 &  - \\
\cline{2-10}
&MANGO   &L-1080  &  89.7 & 89.3 & 85.3 & 94.0 & 93.4 & 88.4 & 9.8 \\
&MANGO   &L-1440 &  90.5 & 90.0 & 86.9 & 94.8 & 94.1 & \textbf{90.1} & 6.3 \\
&MANGO$^*$  &L-1440 &  \textbf{93.4} & \textbf{92.3} & \textbf{88.7} & 92.9 & 92.7 & 88.3 & 6.3 \\
\hline
\hline
\multirow{12}{*}{IC15}
&He et al. $\dagger$  \cite{he2018end} &- &  {82.0} &{ 77.0} & 63.0 & {85.0} & {80.0} & 65.0&  -\\
&FOTS $\dagger$  \cite{liu2018fots} & L-2240  &  81.1 & 75.9 & 60.8 & 84.7 & 79.3 & 63.3 & \textbf{7.5}\\
&TextNet \cite{sun2018textnet} &- &  78.7 & 74.9 & 60.5 & 82.4 & 78.4 & 62.4 & - \\
&Mask TextSpotter$^*$ \cite{liao2019mask}  & S-1600  &  {83.0} & {77.7} & 73.5 & 82.4 & 78.1 & {73.6} &  2.0 \\
&CharNet R-50 \cite{xing2019convolutional} &- &  {83.1} & {79.2} & {69.1} & - & - & - &  - \\
&TextDragon \cite{feng2019textdragon} &- &  {82.5} & {78.3} & {65.2} & 86.2 & 81.6 & {68.0} &  - \\
&Unconstrained \cite{qin2019towards}  &  S-900  &  {83.4} & {79.9} & {68.0} & - & - & - &  - \\
&Boundary \cite{wang2020all} & 1080$\times$1920  &  {79.7} & {75.2} & {64.1} & - & - & - &  - \\
&Text Perceptron \cite{qiao2020text}  & L-2000  &  {80.5} & {76.6} & {65.1} & 84.1 & 79.4 & 67.9 &  - \\
\cline{2-10}
&MANGO  & L-1440  &   80.3 & 77.8 & 66.1 & 84.7 & 81.8 & 69.0 & 6.2\\
&MANGO  & L-1800 &   81.8 & 78.9 & 67.3 &\textbf{ 86.4} & \textbf{83.1} & 70.3 & 4.3\\
&MANGO$^*$  & L-1800  &  \textbf{85.4} & \textbf{80.1} & \textbf{73.9} & 85.2 & 81.1& \textbf{74.6} & 4.3 \\
\hline
\end{tabular}
}
\end{center}
\caption{Results on IC13 and IC15. `S', `W' and `G' mean recognition with strong, weak and generic lexicon, respectively. Superscript `*' means that the method uses the specific lexicons from \cite{liao2019mask}.  Methods marked with $\dagger$ are not support for irregular text. Prefix `L-' and `S-' separately represent that resizing input images by the longer and shorter side.
}
\label{tb:1}
\end{table*}

\subsection{Implementation Details}
All experiments are implemented in Pytorch with 8$\times$32 GB-Tesla-V100 GPUs.

\textbf{Network Details.}
The feature extractor uses ResNet-50 \cite{he2016deep} and FPN \cite{lin2017feature} to obtain fused features from different feature map levels.
Here, the ($4\times$) feature map with $C$$=$$256$ is used to perform the subsequent training and testing tasks.
$L$ is set to 25 to cover most scene text words. The BiLSTM module has 256 hidden units.

\textbf{Training Details.}
All models are trained by the SGD optimizer with batch-size=2, momentum=0.9 and weight-decay=$1\times 10^{-4}$.
In the pretraining stage, the network is trained with an initial learning ratio of $1\times 10^{-2}$ for 10 epochs.
The learning rate is divided by 10 every 3 epochs.
In the finetuning stage, the initial learning rate is set to $1\times 10^{-3}$.
To balance the numbers of synthetic images and real images in each batch, we maintain the sampling ratio of 1:1 for the Curved SynthText dataset versus the other realistic datasets.
The finetuning process lasts for 250k iterations in which the learning rate is divided by 10 at the 120k-th iteration and the 200k-th iteration.

We also conduct the data augmentation for all training processes, including 1) randomly scaling the longer side of the input images to lengths in the range [$720$, $1800$], 2) randomly rotating the images by angles in the range [$-15^\circ, 15^\circ]$, and 3) applying random brightness, jitters, and contrast to input images.

According to the density of text instances in different datasets, we set $S$$=$$60$ for evaluation of IC15 and $S$$=$$40$ for evaluations of IC13, Total-Text and CTW1500.
We simply set all weight parameters as $\lambda_1$$=$$\lambda_2$$=$$\lambda_3$$=$$\lambda$$=$$1$.

\textbf{Testing Details.}
Since the input image's size is an important essential impacting performance, we will report the performance in different input scales, i.e., keep the original ratio and resize the longer side of the image into a fixed value.
All images are tested at a single scale.

Since current implementation only provides rough positions, we modify the end-to-end evaluation metric of \cite{2011End} by considering all detection results with an IoU$>$$0.1$.
In such case, the performance of previous methods will even be decreased due to the decline of precision by some low-grade proposal matching.

\begin{table}
\begin{center}
\scalebox{0.8}{
\begin{tabular}{|l|c|c|c|}
\hline
\multirow{2}{*}{Method}  & \multicolumn{2}{c|}{End-to-End} &\multirow{2}{*}{FPS} \\ \cline{2-3}
  & None & Full & \\
\hline
Mask TextSpotter \cite{liao2019mask} & 65.3 & 77.4 & 2.0\\
CharNet R-50 \cite{xing2019convolutional}   & 66.2 & - &  1.2 \\
TextDragon \cite{feng2019textdragon}  & 48.8 & {74.8} &  - \\
Unconstrained \cite{qin2019towards}  & 67.8 & - &  - \\
Boundary \cite{wang2020all}  & 65.0 & 76.1 &  - \\
Text Perceptron \cite{qiao2020text}   & 69.7 & 78.3 &  - \\
ABCNet \cite{liu2020abcnet}   & 64.2 & 75.7 &  \textbf{17.9} \\
\hline
MANGO (1280)&  {71.7} & 82.6 & 8.9 \\
MANGO (1600)&  \textbf{72.9} & \textbf{83.6} & 4.3 \\
\hline
\end{tabular}
}
\end{center}
\caption{Results on Total-Text. `Full' indicates lexicons of all images are combined. `None' means lexicon-free. The number in brackets is the resized longer side of input image.}
\label{tb:3}
\end{table}
\begin{table}
\begin{center}
\scalebox{0.8}{
\begin{tabular}{|l|c|c|c|}
\hline
\multirow{2}{*}{Method}  & \multicolumn{2}{c|}{End-to-End} & \multirow{2}{*}{FPS}\\ \cline{2-3}
  & None & Full & \\
\hline
Text Perceptron \cite{qiao2020text}   & 57.0 & - & -  \\
ABCNet \cite{liu2020abcnet}   & 45.2 & 74.1 & - \\
\hline
MANGO (1080)&  \textbf{58.9} & \textbf{78.7} & \textbf{8.4} \\
\hline
\end{tabular}
}
\end{center}
\caption{Results on CTW1500. ``Full'' indicates lexicons of all images are combined. ``None'' means lexicon-free. The number in brackets is the resized longer side of input image.}
\label{tb:4}
\end{table}
\subsection{Results on Text Spotting Benchmarks}
\subsubsection{Evaluation of regular text}
We first evaluate our method on IC13 and IC15, following the conventional evaluation metrics \cite{karatzas2015icdar}, two evaluation items (`End-to-End' and `Word Spotting') based on three different lexicons (Strong, Weak, and Generic).

Table \ref{tb:1} shows the evaluation results.
Compared to previous methods evaluated with conventional lexicons, our method achieves the best results on the `Generic' item (except for the end-to-end generic result on IC15), and obtains the competitive results on the rest evaluated items (`Strong' and `Weak').
Compared to the recent state-of-the-art, Mask TextSpotter \cite{liao2019mask} using the specific lexicon, our method obviously outperforms it on all evaluation items.

For the inference speed, though FOTS obtains the highest FPS (Frames Per Second), it fails to handle the irregular cases. Compared with those irregular-based methods, our method achieves the highest FPS.
\begin{figure*}[th!]
\begin{center}
\includegraphics[width=0.98\textwidth, height=4.8cm]{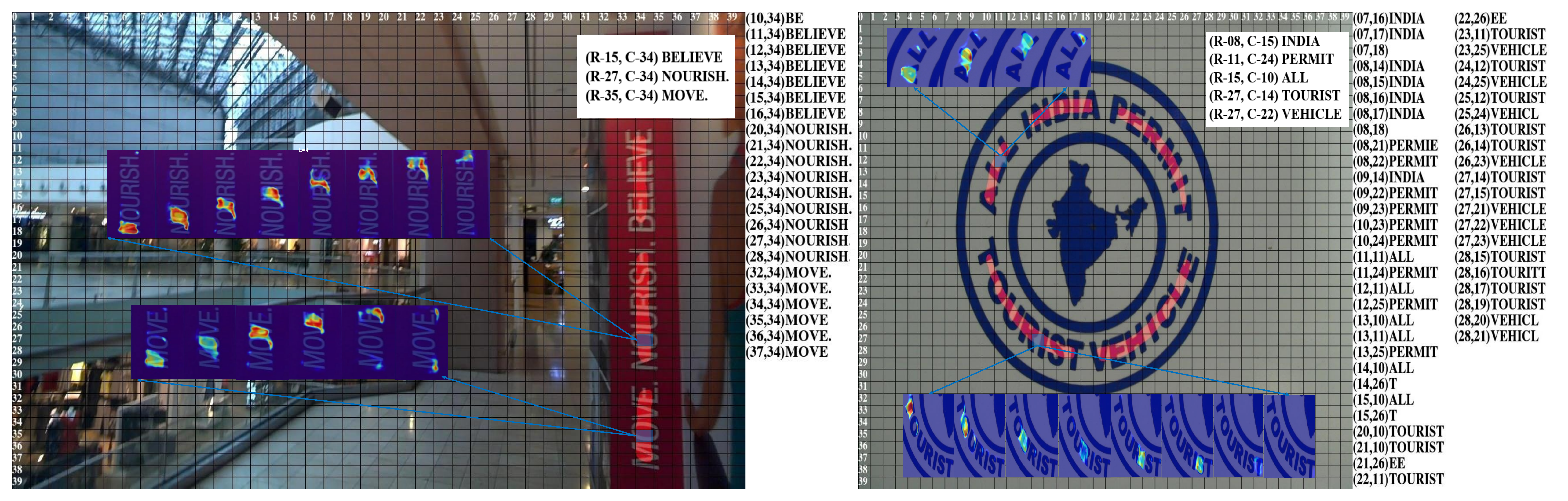}\\
\end{center}
\caption{Visualization of End-to-End recognition on IC15 and Total-Text with some cropped instances' CMA, where $S=40$. The right parts of the images show all the positive predictions before character voting.
Two numbers in the brackets (.,.) separately mean the row and column number.
}
\label{vis1}
\end{figure*}
\begin{figure*}[th!]
\begin{center}
\includegraphics[width=1\textwidth, height=5.8cm]{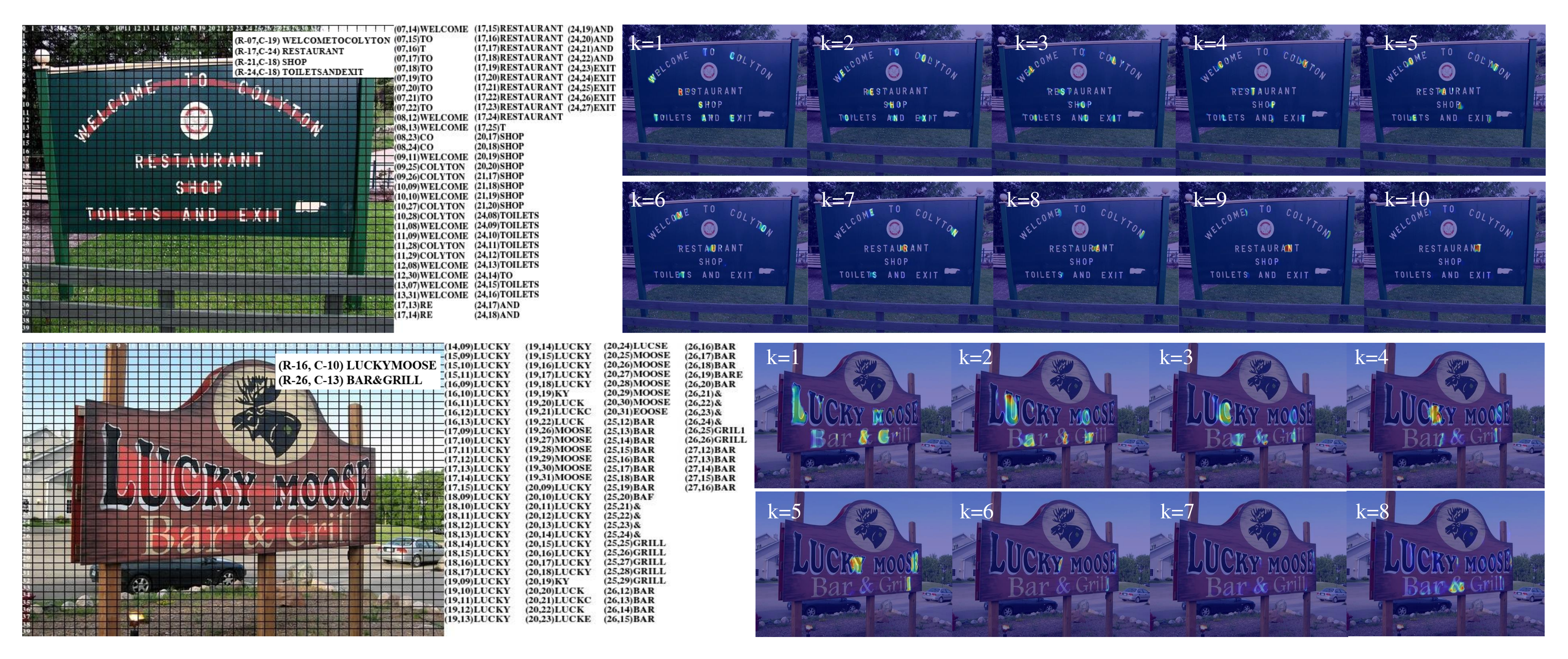}\\
\end{center}
\caption{Visualization of the end-to-end results on SCUT-CTW1500 with the CMA maps in different character positions. 
}
\label{vis3}
\end{figure*}
\begin{table}[th!]
\begin{center}
\scalebox{0.75}{
\renewcommand\tabcolsep{4.0pt}
\begin{tabular}{|c|c|c|c|c|c|c|c|c|c|c|c|}
\hline
\multirow{2}{*}{$S$} & \multicolumn{4}{c|} {IC13 } &  \multicolumn{4}{c|} {IC15 } &  \multicolumn{3}{c|}{Total-Text } \\ \cline{2-12}
& S & W & G &FPS & S & W & G &FPS & None & Full &FPS \\ \cline{2-12}
\hline
20 & 83.2 & 82.5 & 78.7 &\textbf{ 6.58} & 33.8 & 33.0 & 29.1 & \textbf{5.12} &  46.9 & 58.5 & \textbf{4.49} \\
30 & 88.8  & 88.3 & 85.9 & 6.32 & 69.4 & 67.1 & 57.8 & 4.57&  69.8 & 80.6 & 4.37 \\
40 & \textbf{90.5} & \textbf{90.0} & \textbf{86.9} & 6.25 & 80.4 & 77.3 & 66.8 & 4.43 &  72.9 & \textbf{83.6} & 4.28 \\
50 & 90.3 & 89.8 & 86.7 & 6.12 & 81.6 & 78.8 & \textbf{67.8} & 4.38 &  \textbf{73.1} & 83.0 & 4.23 \\
60 & 89.9 & 89.3 & 85.7 & 6.07 & \textbf{81.8} & \textbf{78.9} & 67.3 & 4.27 &  72.2 & 82.9 & 4.21 \\
\hline
\end{tabular}
}
\end{center}
\caption{Evaluation results under different grid numbers.}
\label{tb:S}
\end{table}
\begin{table}[ht]
\begin{center}
\scalebox{0.8}{
\begin{tabular}{|l|c|c|c|c|c|}
\hline
\multirow{2}{*}{Supervision Type}  &  \multicolumn{3}{c|} {IC15 } &  \multicolumn{2}{c|}{Total-Text } \\ \cline{2-6}
& S & W & G & None & Full \\ \cline{2-6}
\hline
Strong & 81.8 & 78.9 & 67.3 & 72.9 & 83.6 \\
Weak & 81.8 & 78.3 & 64.0  & 69.7 & 80.6 \\
\hline
\end{tabular}
}
\end{center}
\caption{Results under different detection supervision types. `Strong' means the original annotations, and `Weak' means rectangular bounding box annotations.}
\label{tb2}
\end{table}
\begin{table*}[h!]
\begin{center}
\scalebox{0.8}{
\begin{tabular}{|l|c|c|c|c|c|c|c|c|}
\hline
Method & Base(100k) & DB & FN & Rotate & Tilt & Weather & Challenge & AP  \\
\hline
SSD300 + HC &  98.3 & 96.6 & \textbf{95.9} & 88.4 & 91.5 & 87.3 & 83.8 & 95.2 \\
RPnet\cite{xu2018towards} &  98.5 & 96.9 & 94.3 & 90.8 & 92.5 & 87.9 & \textbf{85.1} & 95.5 \\
\hline
\hline
MANGO & \textbf{99.0} & \textbf{97.1} & 95.5 & \textbf{95.0} & \textbf{96.5} & \textbf{95.9} & 83.1 & \textbf{96.9} \\
\hline
\end{tabular}
}
\end{center}
\caption{End-to-End recognition precision results on CCPD.}
\label{ccpd}
\end{table*}
\subsubsection{Evaluation of irregular text}
We test our method on Total-Text, as shown in Table \ref{tb:3}.
We see that our method surpasses the state-of-the-art by 3.2\% and 5.3\% in ``None" and ``Full" metrics.
Notice that even without an explicit rectification mechanism, our model can handle irregular text well only driven by the recognition supervision.
Though the inference speed is about 1/2 of ABCNet with the test scale of 1280, our method achieves the remarkable performance gains.

We also evaluate our method on CTW1500. There are few works that reported the end-to-end results because it mainly contains the line-level text annotations.
To adapt to this situation, we retrain the detection branch on the training set of CTW1500 to learn the line-level centerline segmentation, and fix the weights of the backbone and other branches. Note that the recognition will not be affected and still output the word-level sequences. The final results will be simply concatenated from left to right according to the inferred connected regions. Chinese characters are set as NOT CARE.

Results are shown in Table \ref{tb:4}. We find that our method obviously surpasses previous advances by 1.9\% and 4.6\% on `None' and `Full' metrics, respectively.
Therefore, we believe that if there are enough data with only line-level annotations, our model can adapt to such scenarios well.

\subsection{Visualization Analysis}
Figure \ref{vis1} visualizes the end-to-end text spotting results on IC15 and Total-Text.
We detailedly show the prediction results of each positive grids ($o_{i,j}$$>$$0.3$) before character voting.
We see that our model can correctly focus on the corresponding position and learn the complex reading order of character sequences for arbitrary-shaped (\emph{e.g.} curved or vertical) text instances.
After the character voting strategy, the word with the highest confidence will be generated.


We also demonstrate some of the results of CTW1500 with their visualized CMA, as shown in Figure \ref{vis3}.
Note that we only fine-tune the line-level segmentations part based on the dataset's position labels while fixing the remaining parts.
Here, We visualize the feature maps of CMAs by overlaying all grids' attention map in the same character position ($k$) as:
\begin{equation}
x^{*}_{char}[k]=\sum_{i \in S^2} x_{char}[i][k]
\end{equation}
where $x^{*}_{char}[k]\in \mathbb{R}^{L\times H \times W}$, and $k=1,2,...,L$.
As shown in Figure \ref{vis3}, we see that model indeed pays attention to all correct character positions of all text instances in the image at the same time.  At the end of each text instance, there is a highlight region that means the `EOS' position's attention.


\subsection{Ablation Studies}
\subsubsection{Ablation of grid numbers}
The grid number $S^2$ is a crucial parameter affecting the final results.
If $S$ is too small, there will be too many texts occupying the same grid. Otherwise, too big of $S$ will result in more computation cost.
Here, we conduct experiments to find the feasible value of $S$ for different datasets.

From Table \ref{tb:S}, we find that the best $S$ for both IC13 and Total-Text is $40$.
The value for IC15 is $60$. This is because IC15 contains more dense and small instances.
In sum, the overall performance increases along with increasing of $S$ and becomes stable when $S$$>$$=40$.
Of course, FPS will decrease slightly along with increasing of $S$.

\subsubsection{Evaluation of Coarse Position Supervision}
As mentioned above, our method can be learned well with only rough position information.
To demonstrate this, we also conduct the experiments to transfer all localization annotations as the form of rectangular bounding boxes.  We simply adopt the RPN head as the detection branch.

Table \ref{tb2} shows the results on IC15 and Total-Text.
Even with the rough position supervision, MANGO only decreases the performance ranging from 0\% to 3\%, and is comparable with the state-of-the-arts.
Note that, the coarse position only serves the grid selection so that it can be simplified as much as possible according to specific tasks' requirement.

\subsection{Challenging License Plate Recognition without Position Annotations}
To demonstrate the model's generalization ability, we conduct experiments to evaluate the end-to-end license plate recognition results on a public dataset, CCPD \cite{xu2018towards}.
For fairness, we follow the same experimental settings and use the initially released version of the dataset with 250k images.
The CCPD-Base dataset is separated into two equal parts: 100k samples for training and 100k samples for testing.
There are six complex testing sets (including DB, FN, Rotate, Tilt, Weather, and Challenge) for evaluating the algorithm's robustness, which have 50k images in total.

Since each image in CCPD contains only one plate, our model can be further simplified by removing the detection branch to  predict the final character sequence directly. Therefore, the grid number is reduced to $S=1$, and the maximum sequence length is set to $L=8$.
We directly fine-tune the model (having been pre-trained by SynthText) on CCPD training set with only the sequence-level annotations, and then evaluate the final recognition accuracy on the above seven testing datasets. The testing phase is performed on the original image with a size of $720\times 1160$.
\begin{figure}[!htpt]
\begin{center}
\includegraphics[width=0.9\linewidth,height=2.6cm]{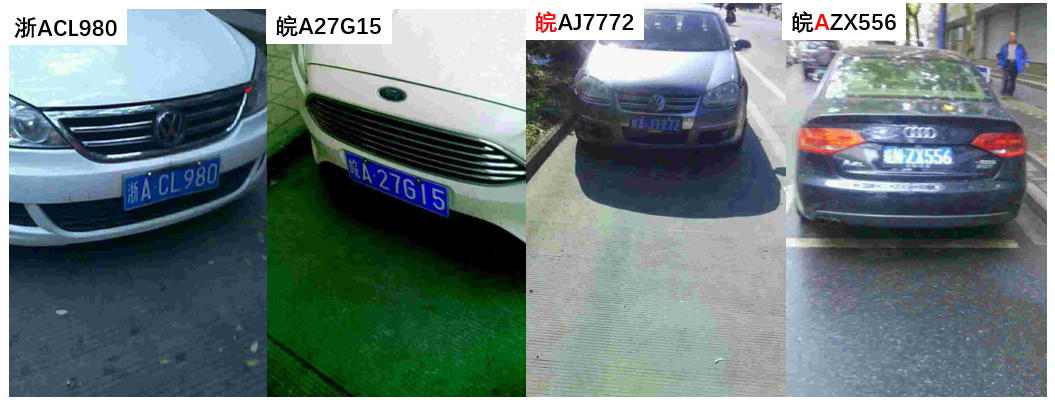}\\
\end{center}
   \caption{
   Visualization result on CCPD. Since $S=1$, no position information is involved.
   }
\label{fig:ccpd}
\end{figure}

Table \ref{ccpd} shows the end-to-end recognition results. 
Although the proposed method is not designed for the license plate recognition task, it still can be easily transferred to such scenarios.
We see that the proposed model outperforms previous methods in 5 out of 7 test sets and achieves the highest average precision.
Figure \ref{fig:ccpd} shows some visualization results on the CCPD test sets.
The failure samples are mainly from the situation that images are too blurred to be recognized.

This experiment demonstrates that in many situations with only one text instance (\emph{e.g.,} industrial printing recognition or meter dial recognition), a good End-to-End model can be obtained without detection annotations.

\section{Conclusion}
In this paper, we propose a novel one-staged scene text spotter named {MANGO}.
This model removes the RoI operations and designs the position-aware attention module to coarsely localize the text sequences.
After that, a lightweight sequence decoder is applied to obtain all of the final character sequences into a batch.
Experiments show that our method achieves competitive and even state-of-the-art results on popular benchmarks.

\newpage
{\footnotesize{
\bibliographystyle{aaai}
\bibliography{reference}
}
}
\end{document}